\documentclass[conference]{IEEEtran}
\usepackage{times}

\usepackage[numbers]{natbib}
\usepackage{multicol}
\usepackage[colorlinks=true,bookmarks=true]{hyperref}

\usepackage{amsmath}
\usepackage{balance}
\usepackage{amssymb}
\usepackage{bbm}
\usepackage{dsfont}
\usepackage[table]{xcolor}
\usepackage[font=scriptsize]{caption}
\usepackage{subcaption}
\usepackage{graphicx}
\usepackage{xcolor}
\usepackage{pifont}
\newcommand{\xmark}{\ding{55}}%
\usepackage{booktabs}
\usepackage{multirow}
\usepackage{graphicx}
\usepackage{float}
\title{\Large {\bf
ShaSTA}: Modeling Shape and Spatio-Temporal Affinities for 3D Multi-Object Tracking}

\begin{document}

\author{\authorblockN{Tara Sadjadpour}
\authorblockA{Stanford University\\
Stanford, CA 94305\\
tsadja@stanford.edu}
\and
\authorblockN{Jie Li}
\authorblockA{NVIDIA\\
Santa Clara, CA 95051\\
jieli@nvidia.com}
\and
\authorblockN{Rares Ambrus}
\authorblockA{Toyota Research Institute\\
Los Altos, CA 94022\\
rares.ambrus@tri.global}
\and
\authorblockN{Jeannette Bohg}
\authorblockA{Stanford University\\
Stanford, CA 94305\\
bohg@stanford.edu}}


%

\makeatletter
\g@addto@macro\@maketitle{
    \begin{figure}[H]
    \begin{minipage}{\textwidth}
    \includegraphics[width=\textwidth]{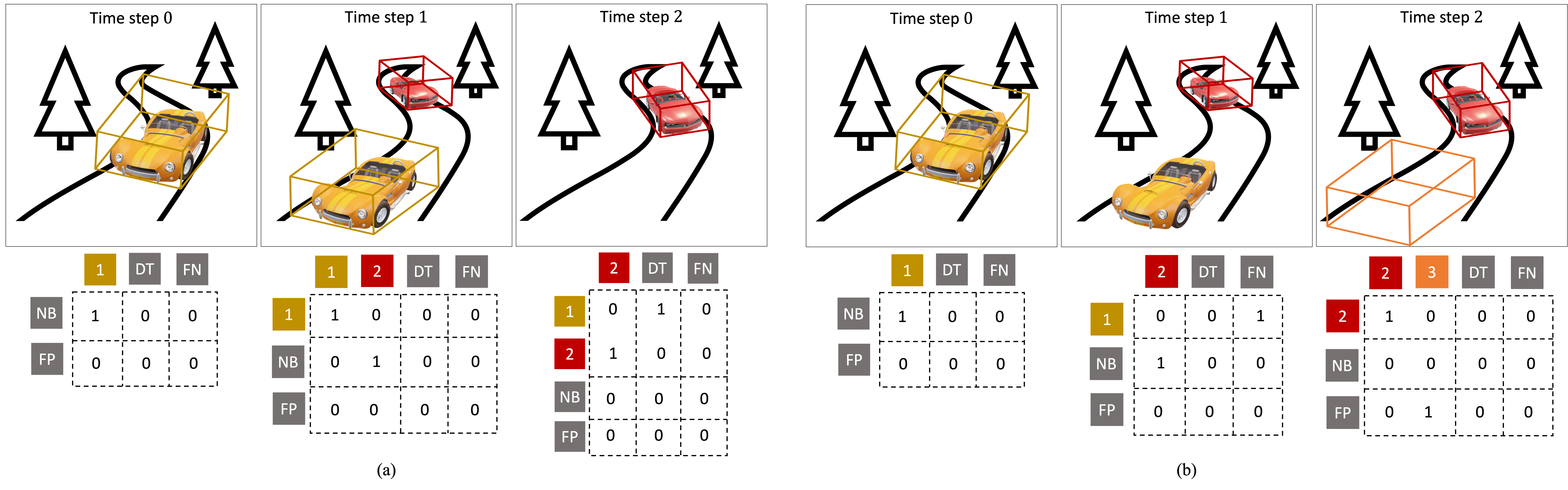}
    \captionof{figure}{
        Examples for tracking scenarios (top) and associated ground-truth affinity matrices (bottom) that our algorithm, ShaSTA, must estimate. In all affinity matrices, the rows correlate with previous frame tracks that are represented with single-history bounding box detections, while the columns correlate with current frame detections. The affinity matrices are augmented with two rows for {\em newborn track\/} (NB) and {\em false-positive\/} (FP) anchors that the current frame detections can match with, and two columns for {\em dead track\/} (DT) and {\em false-negative\/} (FN) anchors that the previous frame tracks can match with. These four augmentations are learned representations that capture the essence of these four detection and track types for each frame. The bounding box color corresponds to the track ID. At time step $t=0$, both figures (a) and (b) start with affinity matrix $A_0$, where the yellow car is detected and is matched with NB, thus initializing Track 1. For Figure (a), at $t=1$, $A_1$ shows that the yellow car is detected again and matched with the previous frame's Track 1, while the red car is detected as a newborn with Track ID 2. Then, at $t=2$, there is only one detection for the red car, so it is matched with the previous frame's Track 2, while Track 1 is matched with DT in $A_2$. In Figure (b), at time step $t=1$, the red car is detected as a newborn with Track ID 2, while the yellow car is not detected at all so Track 1 from $t=1$ is matched with FN. Finally, at $t=2$, the red car is detected again and matched with Track 2, while a detection is found in a region with no object so it is matched with FP in $A_2$.
    }\label{fig:teaser}%
    \end{minipage}
    \end{figure}
    \vspace{-16pt}
}
\makeatother

\maketitle
\renewcommand\thefigure{\arabic{figure}}
\setcounter{figure}{1}

\begin{abstract}


Multi-object tracking (MOT) is a cornerstone capability of any robotic system. The quality of tracking is largely dependent on the quality of the detector used. In many applications, such as autonomous vehicles, it is preferable to over-detect objects to avoid catastrophic outcomes due to missed detections. As a result, current state-of-the-art 3D detectors produce high rates of false-positives to ensure a low number of false-negatives. This can negatively affect tracking by making data association and track lifecycle management more challenging. Additionally, occasional false-negative detections due to difficult scenarios like occlusions can harm tracking performance. To address these issues in a unified framework, we propose to learn shape and spatio-temporal affinities between tracks and detections in consecutive frames. Our affinity provides a probabilistic matching that leads to robust data association, track lifecycle management, false-positive elimination, false-negative propagation, and sequential track confidence refinement. Though past 3D MOT approaches address a subset of components in this problem domain, we offer the first self-contained framework that addresses all these aspects of the 3D MOT problem. We quantitatively evaluate our method on the nuScenes tracking benchmark where we achieve 1st place amongst LiDAR-only trackers using CenterPoint detections. Our method estimates accurate and precise tracks, while decreasing the overall number of false-positive and false-negative tracks and increasing the number of true-positive tracks. Unlike past works, we analyze our performance with 5 metrics, including AMOTA, the most common tracking accuracy metric. Thereby, we give a comprehensive overview of our approach to indicate how our tracking framework may impact the ultimate goal of an autonomous mobile agent. We also present ablative experiments, as well as qualitative results that demonstrate our framework's capabilities in complex scenarios. Please see our project website for source code and demo videos: \href{https://sites.google.com/view/shasta-3d-mot/home}{sites.google.com/view/shasta-3d-mot/home}.

\end{abstract}
\IEEEpeerreviewmaketitle

\section{Introduction}
In this work, we address the problem of 3D multi-object tracking using LiDAR point cloud data as input. 3D multi-object tracking is an essential capability for autonomous agents to navigate effectively in their environments. For example, autonomous cars need to understand the motion of surrounding traffic agents to drive  safely. Online tracking is achieved by (1) accurately matching uncertain detections to existing tracks and (2) determining when to birth and kill tracks of unmatched observations. These two processes are referred to as {\em data association\/} and {\em track lifecycle management\/}, respectively. 

Most recent approaches~\cite{weng2019ab3dmot, chiu2020probabilistic, chiu2021probabilistic, stearns2022spot, meyer2018message, liang2022neural, yin2021center, zaech2022learnable, kim2021eagermot, wang2022deepfusionmot, weng2020gnn3dmot} toward 3D multi-object tracking in autonomous driving apply the tracking-by-detection paradigm whereby off-the-shelf detections are available from a state-of-the-art 3D detector. Though this framework has been successful, there exists a practical gap between the objectives of 3D detection and tracking. Since missing a detection can lead to catastrophic events in self-driving, 3D detectors are evaluated on the full recall spectrum of detections (e.g., mAP~\cite{everingham2010pascal}) and are deployed with the goal of having high recall. As a result, they tend to generate high rates of false-positives to ensure a low number of false-negatives~\cite{yin2021center, zhou2018voxelnet}. These redundant detections lead to highly cluttered outputs that make data association and track lifecycle management more challenging. Additionally, occasional false-negative detections due to factors such as heavy occlusion can cause significant fragmentations in the tracks or lead to the premature termination of tracks. 

In this work, we address the task of LiDAR-only tracking. In contrast to camera-only and camera-fusion tracking~\cite{chen2022polar,weng2020gnn3dmot,chiu2021probabilistic,kim2021eagermot, wang2022deepfusionmot}, LiDAR tracking offers many practical advantages, balancing high accuracy with a less complex system configuration and lower computational demands. Still, LiDAR-only systems face significant challenges due to the sparse, unstructured nature of LiDAR point clouds. Recent works in LiDAR-only tracking-by-detection~\cite{stearns2022spot, meyer2018message, liang2022neural, zaech2022learnable} have attempted to develop learning-based and statistical approaches to better handle data association, false-positive elimination, false-negatives propagation, and track lifecycle management.  However, these approaches do not effectively leverage spatio-temporal and shape information from the LiDAR sensor input, and in some cases, only rely on detected bounding box parameters for spatio-temporal information.

Our {\bf main contribution\/} is a 3D multi-object tracking framework called {\em ShaSTA\/} that models \textbf{sha}pe and \textbf{s}patio-\textbf{t}emporal \textbf{a}ffinities between tracks and detections in consecutive frames. By better understanding objects' shapes and spatio-temporal contexts, \textit{ShaSTA} improves data association, false-positive (FP) elimination, false-negative (FN) propagation, newborn (NB) initialization, dead track (DT) termination, and track confidence refinement. 
At its core, \textit{ShaSTA} learns an affinity matrix that encodes shape and spatio-temporal information to obtain a probabilistic matching between detections and tracks. {\em ShaSTA\/} also learns representations of FP, NB, FN, and DT \textit{anchors} for each frame. By computing the affinity between detections and tracks to these learned representations, we can classify detections as newborn tracks or false-positives, while tracks can be classified as dead tracks or false-negatives. Once we use these matches to form tracks, {\em ShaSTA\/} continues to leverage the affinity estimation with a novel sequential track confidence refinement technique that represents track score refinement as a time-dependent problem and leverages information extracted from shape and spatio-temporal encodings to obtain significant improvements in overall tracking accuracy.

This overall approach not only aids in tracking performance but also addresses concerns for real-world deployment and downstream decision-making tasks by eliminating false-positive tracks and minimizing false-negative tracks. {\em ShaSTA\/} offers an efficient and practical implementation that can re-use the LiDAR backbone of most off-the-shelf LiDAR-based 3D detectors. Without loss of generality, we choose the LiDAR backbone from CenterPoint \cite{yin2021center} for a fair comparison against most previous works. We demonstrate that our approach achieves state-of-the-art results on the nuScenes tracking benchmark~\cite{caesar2020nuscenes} where we achieve significant gains in overall tracking accuracy and precision, while improving the raw number of true-positive, false-positive, and false-negative tracks. We also include ablation studies to analyze the contribution of each component.

\section{Related Work}\label{sec:related}

\subsection{LiDAR-based 3D Detection}

Most recent 3D multi-object tracking (3D MOT) works follow the tracking-by-detection paradigm~\cite{weng2019ab3dmot, liang2020pnpnet, chiu2021probabilistic, zaech2022learnable, wang2022deepfusionmot, meyer2018message}, making the quality of 3D detectors a key component to tracking performance~\cite{chiu2021probabilistic, luiten2021hota, wang2022camo}.
LiDAR-based 3D detectors extend the image-based 2D detection algorithms~\cite{girshick2015fast, zhou2019objects} to 3D.
The typical detector starts with an encoder~\cite{qi2017pointnet,engelcke2017vote3deep,yang2018pixor} applied to unstructured point cloud data to extract an intermediate 3D feature grid~\cite{zhou2018voxelnet} or bird's-eye-view (BEV) feature map~\cite{lang2019pointpillars,yin2021center}. Then, a decoder is applied to extract objectness~\cite{choi2019objectness} and detailed object attributes~\cite{qi2019deep,yin2021center}.
Our approach proposes to leverage the intermediate BEV feature map for robust multi-object tracking, which can be generalized to most modern 3D detectors. CenterPoint detections have become a standardized 3D detection in the 3D MOT community, since the authors of the work released the official detection files for all data splits. As of this writing, CenterPoint has the highest mAP of all publicly-released 3D detections. Thus, in this work, we apply CenterPoint~\cite{yin2021center} for fair comparison against existing 3D MOT techniques.

\subsection{3D Multi-Object Tracking}

With the advances in 3D detection, the 3D MOT community has been focusing on four directions to establish robust tracking within the tracking-by-detection paradigm: motion prediction, data association, track lifecycle management, and confidence refinement.

In 3D MOT, abstracted detection predictions are often used for motion prediction and data association, which is completed using pair-wise feature distance between tracks and detections~\cite{weng2019ab3dmot, chiu2020probabilistic, yin2021center}. AB3DMOT~\cite{weng2019ab3dmot} provides a baseline to combine Kalman Filters and Intersection Over Union (IoU) association for 3D MOT. Subsequent works in this line explore variations of data association metrics, such as the Mahalanobis distance to leverage uncertainty measurements in the Kalman Filter~\cite{chiu2020probabilistic}. As an alternative, CenterPoint~\cite{yin2021center} uses high-fidelity instance velocity predictions as a motion model to complete data association. Unlike these works that only use low-dimensional bounding boxes and explicit motion models for tracking, we propose to leverage intermediate representations from a detection network for richer information about each object, while also modeling the global relationship of all the objects in a scene to improve data association. 



Additionally, track lifecycle management has been reasonably successful with heuristics that are tuned to each dataset~\cite{weng2019ab3dmot,chiu2020probabilistic,yin2021center}, but these hand-tuned techniques have shortcomings in more complex situations. Recently, OGR3MOT~\cite{zaech2022learnable} has proposed combining tracking and predictive modeling through a unified graph representation that learns data associations and track lifecycle management, FP elimination, and FN interpolation. Additionally,~\cite{meyer2018message} proposes a statistical approach for data association under a Bayesian estimation framework. This work was later extended with Neural-Enhanced Belief Propagation (NEBP)~\cite{liang2022neural} to include a learning-based aspect to the framework for FP suppression. Though these techniques yield competitive results compared to heuristics, their main shortfall is their inability to capture object shape information. Unlike these techniques, our approach leverages raw LiDAR data to capture shape information about objects in our scene, which we can use to further improve tracking performance. 

Finally, track confidence is a less-explored but also important aspect of the tracking problem. Track confidence indicates the quality of a track relative to tracks for the same object class within a given frame. Just as it is the case for detection confidences, the most important characteristic of a track's confidence is its value relative to other tracks' confidences. 
In most 3D MOT works, algorithms assign the track confidence to be the same as that of the confidence score that comes with off-the-shelf detections matched to the track~\cite{weng2019ab3dmot,chiu2020probabilistic,yin2021center,zaech2022learnable}. In a recent effort to improve on this aspect of the tracking problem,~\cite{stearns2022spot} used an existing detection score refinement technique from~\cite{simonelli2019disentangling} to obtain new tracking scores and improve the overall tracking performance. Though this method demonstrates improvements in ablative studies, its main weakness is that it directly applies a detection score refinement technique to the track score, overlooking the fact that unlike detections, tracks are ever-changing, time-dependent quantities that cannot be treated in isolation for the current time step. In this paper, we argue that track score refinement should be a sequential process that reflects changing environmental factors. Thus, we propose a first-of-its-kind approach that represents track score refinement as a sequential problem and leverages shape and spatio-temporal information encoded in our learned affinity matrix to obtain significant improvement in overall tracking accuracy.

\begin{figure*}[ht!]
\centering
\includegraphics[width=\linewidth]{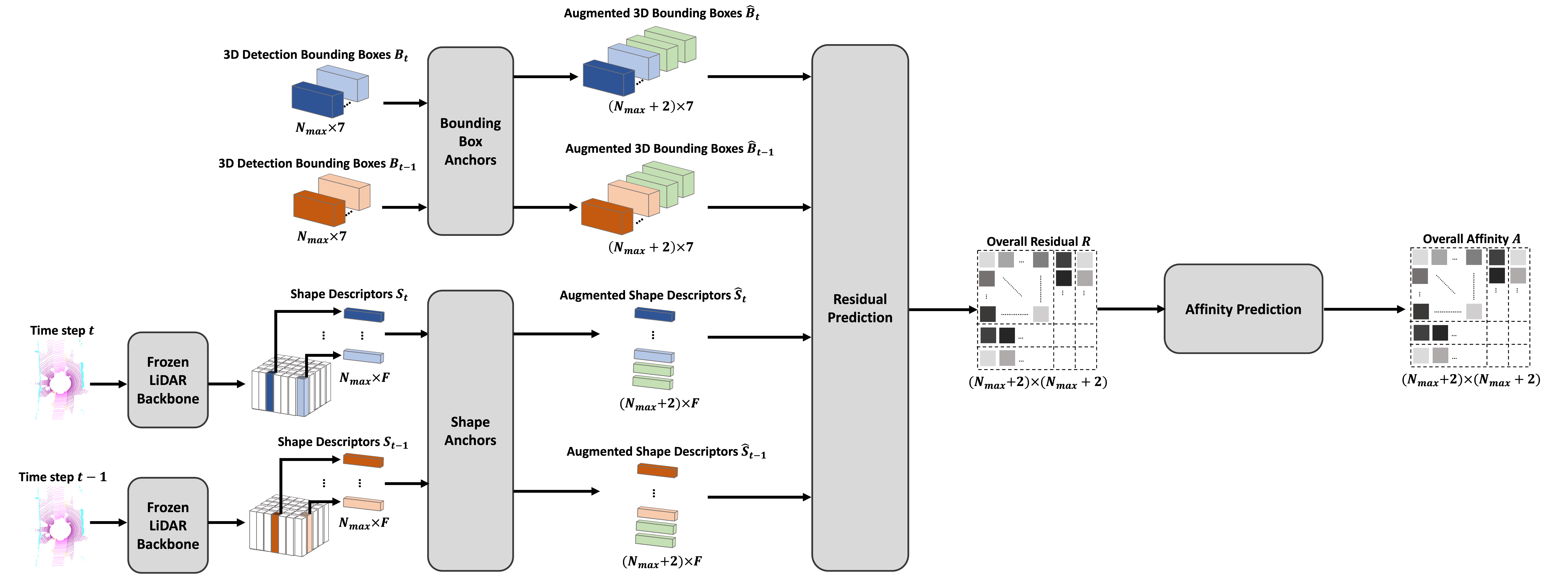}
\caption{Algorithm Flowchart. Going from top to bottom and left to right, ShaSTA uses $N_{max}$ low-dimensional bounding boxes from the current and previous frames to learn bounding box representations for the FP, NB, FN, and DT anchors. The FP and NB anchors are appended to previous frame bounding boxes to form the augmented bounding boxes $\hat{B}_{t-1}$, while the FN and DT anchors are added to the current frame detections to form the augmented bounding boxes $\hat{B}_{t}$. Using the pre-trained frozen LiDAR backbone from our off-the-shelf detector, we extract shape descriptors that are then used to learn shape representations for our anchors. The learned shape representations for the anchors are appended to create augmented shape descriptors. These augmented bounding boxes and shape descriptors are then used to find a residual that captures the spatio-temporal and shape similarities between current and previous frame detections, as well as between detections and anchors. The residual is then used to predict the affinity matrix for probabilistic data associations, track lifecycle management, FP elimination, FN propagation, and track confidence refinement.}
\label{fig:pipeline}
\end{figure*}

Another line of 3D MOT algorithms focuses on using camera-LiDAR fusion to learn data association, track lifecycle management, and FP suppression~\cite{weng2020gnn3dmot,chiu2021probabilistic,kim2021eagermot, wang2022deepfusionmot}. However, in this work we focus on LiDAR-only 3D MOT due to the practical advantages it offers. First, it is more cost-efficient to use fewer sensors. Additionally, LiDAR-only tracking reduces the risk of calibration and synchronization issues between various sensor modalities in a more complex system configuration.

\section{SHASTA}
Our algorithm operates under the tracking-by-detection paradigm and is visualized in Fig.~\ref{fig:pipeline}. ShaSTA extracts shape and spatio-temporal relationships between consecutive frames to learn affinity matrices that capture associations between detections and tracks. ShaSTA learns bounding box and shape representations for FP, NB, FN, and DT {\em anchors\/}, where each anchor captures the common features between the set of detections and tracks that falls under each of these four categories in a given frame. Thus, ShaSTA's affinity matrix not only estimates the probabilities of previous frame tracks matching with current frame detections, but also considers previous frame tracks matching with DT or FN anchors and current frame detections matching with NB or FP anchors. Unlike past techniques that only use low-dimensional bounding box representations, ShaSTA leverages shape and spatio-temporal information from the LiDAR sensor input, resulting in a robust data association technique that effectively accounts for FP elimination, FN propagation, NB initialization, and DT termination.

\subsection{Affinity Matrix Usage for Online Tracking}
Allowing for up to $N_{max}$ detections per frame, ShaSTA estimates the affinity matrix $A \in \mathbb{R}^{(N_{max}+2) \times (N_{max}+2)}$. As shown in Figure~\ref{fig:teaser}, the rows of our affinity matrix represent the previous frame tracks and the columns represent our current frame detections. The two augmented columns are for the DT and FN anchors, so previous frame tracks can match with DT or FN. Similarly, because current frame detections can match with NB or FP, the affinity matrix has two augmented rows for the NB and FP anchors. Though there can be no more than one match for each bounding box in the current and previous frames, the same is not true for our learned anchors, i.e. more than one current frame bounding box can match with the NB or FP anchors, respectively. To handle this situation, we create a forward matching affinity matrix $A_{fm} \in \mathbb{R}^{N_{max}\times (N_{max}+2)}$ by removing the two row augmentations from $A$ and applying a row-wise softmax so that more than one previous frame track can have sufficient probability to match with the DT and FN anchors, respectively: 
\begin{align}
    A_{fm} =  \text{softmax}_\text{row}\left(A_1\right).
\end{align}

Using similar logic, we create a backward matching affinity matrix $A_{bm} \in \mathbb{R}^{(N_{max}+2) \times N_{max}}$ by removing the two column augmentations from $A$ and applying a column-wise softmax so that more than one current frame detection can have sufficient probability to match with the NB and FP anchors, respectively: 
\begin{align}
    A_{bm} =  \text{softmax}_\text{col}\left(A_2\right).
\end{align}

Here, forward matching indicates that we want to find the best current frame match for each previous frame track, while backward matching refers to finding the best previous frame match for each current frame detection.


To form tracks, we combine the affinity matrix outputs and the matching algorithm from \cite{yin2021center} as follows. We take $A_{fm}$ and label any previous frame track that has a probability above the threshold $\tau_{dt}$ in the DT column as a DT, and we forward propagate any previous frame track into the current frame if it has a probability above the threshold $\tau_{fn}$ in the FN column. Therefore, in addition to provided detections, {\em forward propagation\/} creates a new detection for the current frame to handle occluded objects that off-the-shelf detections missed. We create a box in the current frame $b'_t := (x',y',z,w,l,h,r_y)$, by moving the 2D center coordinate of the existing track $b_{t-1} := (x,y,z,w,l,h,r_y)$ to the next frame based on the estimated velocity provided by our off-the-shelf 3D detector and $\Delta t$ between the two frames, i.e. $x' = x + v_x\Delta t$ and $y' = y + v_y\Delta t$. All previous frame tracks that do not qualify as FN or DT are kept as is. Analogously, we take $A_{bm}$ and remove any current frame detection that has a probability above the threshold $\tau_{fp}$ in the FP row and we label any current frame detection that has a value above the threshold $\tau_{nb}$ in the NB row as an NB. All current frame detections that do not qualify as FP or NB are kept as is. See Section~\ref{sec:training} for details on the threshold values we choose.

With these detections, we then run the greedy algorithm from \cite{yin2021center}. In its original form, the greedy algorithm takes all unmatched detections and initializes NBs with them. However, we check if an unmatched detection is labeled as an NB in the affinity matrix based on the threshold $\tau_{nb}$ and whether it is outside the maximum distance threshold with respect to other tracks to initialize it as an NB. Otherwise, it is discarded. Additionally, we check if an unmatched track is labeled as a DT based on the threshold $\tau_{dt}$ and whether it is outside the maximum distance threshold with respect to other detections to terminate it as a DT.


\subsection{Learning to Predict Affinity Matrices}
ShaSTA uses off-the-shelf 3D detections, as well as the pre-trained LiDAR backbone used to generate the detections. ShaSTA first learns bounding box and shape representations for the anchors so we can later use this information to learn a residual that encodes spatio-temporal and shape similarities not only between current frame detections and previous frame tracks, but also between current frame detections and FP and NB anchors or between previous frame tracks and FN and DT anchors. The residual is then used to predict affinity matrices that assign probabilities for matching detections to tracks, eliminating FPs, propagating FNs, initializing NBs, and terminating DTs. 

\subsubsection{Learning Bounding Box Representations for Anchors} 
Our goal is to learn bounding box representations for each of the FP, FN, NB, and DT anchors in a given frame pair. The aim of the bounding box representation of each anchor is to capture the commonalities between detection bounding boxes that fall under each of these four categories. 

We use off-the-shelf 3D detections, which provide us with 3D bounding box representations that include each box's center coordinate, dimensions, and yaw rotation angle: $b := (x,y,z,w,l,h,r_y)$. We take the set of bounding boxes for the current frame $B_t \in \mathbb{R}^{N_{max} \times 7}$ and previous frame $B_{t-1} \in \mathbb{R}^{N_{max} \times 7}$. We fix the maximum number of bounding boxes we take per frame to be $N_{max}$; we zero pad the matrix if there are fewer than $N_{max}$ bounding boxes, and we sample the top $N_{max}$ detection bounding boxes if there are greater than $N_{max}$ boxes. 

Then, we create a learned bounding box representation at time step $t$ for FP and NB anchors using the current frame detections $B_t$ as follows, where each $\sigma$ represents an MLP: 
\begin{align}
    b_{fp} &= \sigma_{fp}^b(B_t) \\
    b_{nb} &= \sigma_{nb}^b(B_t).
\end{align}

Similarly, we find learned bounding box representations for FN and DT anchors using previous frame detections $B_{t-1}$: 
\begin{align}
    b_{fn} &= \sigma_{fn}^b(B_{t-1}) \\
    b_{dt} &= \sigma_{dt}^b(B_{t-1}).
\end{align}

For all four cases, we apply the absolute value to the MLP outputs corresponding to $(w,l,h)$, since dimensions need to be non-negative values. We then concatenate $b_{fp} \in \mathbb{R}^7$ and $b_{nb} \in \mathbb{R}^7$ to $B_{t-1}$ to get $\hat{B}_{t-1} \in \mathbb{R}^{(N_{max}+2) \times 7}$, as well as $b_{fn} \in \mathbb{R}^7$ and $b_{dt} \in \mathbb{R}^7$ to $B_{t}$ to get $\hat{B}_{t} \in \mathbb{R}^{(N_{max}+2) \times 7}$. The reason we append FP and NB anchors to $B_{t-1}$ is so that current frame detections can match to them, and the same logic applies for appending FN and DT anchors to $B_t$.

\subsubsection{Learning Shape Representations for Anchors} 
In this section, we extract shape descriptors for each 3D detection using the pre-trained LiDAR backbone of our off-the-shelf detector to leverage spatio-temporal and shape information from the raw LiDAR data. In similar fashion to the previous section, our goal is to learn shape representations for the FP, FN, NB, and DT anchors using the shape descriptors from existing detections. These shape descriptors will be used to match detections to each other or one of the anchors if they share similar shape information.

We pass the current frame's 4D LiDAR point cloud with an added temporal dimension into the frozen pre-trained VoxelNet \cite{zhou2018voxelnet} LiDAR backbone used in the CenterPoint~\cite{yin2021center} detector. This outputs a bird's-eye-view (BEV) map for the current frame, where each voxelized region encodes a high-dimensional volumetric representation of the shape information in that region. Then, for each current frame detection bounding box, we extract a shape descriptor using bilinear interpolation from the BEV map as in \cite{yin2021center} using the bounding box center, left face center, right face center, front face center, and back face center. We concatenate these 5 shape features to create the overall shape descriptor for each bounding box. Note that in BEV, the box center, bottom face center, and top face center all project to the same point, so we forgo extracting the latter two centers. We accumulate all of the shape features extracted for the current frame detections and call this cumulative shape descriptor $S_t \in \mathbb{R}^{N_{max}\times F}$. We repeat the same procedure for the previous frame's LiDAR point cloud and bounding boxes to get the overall shape descriptor $S_{t-1} \in \mathbb{R}^{N_{max}\times F}$. 

Using the extracted shape features for the current frame $S_t$, we obtain our learned shape descriptors for the FP and NB anchors with MLPs $\sigma_{fp}$ and $\sigma_{nb}$, respectively: 
\begin{align}
    s_{fp} &= \sigma_{fp}^s(S_t) \\
    s_{nb} &= \sigma_{nb}^s(S_t).
\end{align}

Likewise, we use our previous frame's shape features $S_{t-1}$ to learn shape descriptors for FN and DT anchors as such: 
\begin{align}
    s_{fn} &= \sigma_{fn}^s(S_{t-1}) \\
    s_{dt} &= \sigma_{dt}^s(S_{t-1}).
\end{align}

Just as we did in the previous section, we concatenate $s_{fp} \in \mathbb{R}^{F}$ and $s_{nb} \in \mathbb{R}^{F}$ to $S_{t-1}$ to get the augmented shape descriptor $\hat{S}_{t-1} \in \mathbb{R}^{(N_{max}+2)\times F}$, as well as $s_{fn} \in \mathbb{R}^{F}$ and $s_{dt} \in \mathbb{R}^{F}$ to $S_{t}$ to get $\hat{S}_t \in \mathbb{R}^{(N_{max}+2)\times F}$.

\subsubsection{Residual Prediction} 
Using the augmented 3D bounding boxes and shape descriptors, we aim to find three residuals that measure the similarities between current and previous frames' bounding box and shape representations. Since the boxes are low-dimensional abstractions, we aim to maximize the amount of spatio-temporal information we extract from them. Thus, we obtain two residuals for the augmented 3D bounding boxes called the VoxelNet and bounding box residuals, $R_v$ and $R_b$ respectively. We also learn one shape residual $R_s$ between the augmented shape descriptors. We obtain our overall residual $R$ that captures the spatio-temporal and shape similarities by taking a weighted sum of the three residuals, where the weights are also learned. 

\noindent \textbf{VoxelNet Residual.} Our first residual $R_v \in \mathbb{R}^{({N_{max}+2}) \times ({N_{max}+2})}$ is a variation of the VoxelNet \cite{zhou2018voxelnet} bounding box residual. We adapt the VoxelNet residual for the end-goal of data assocation between $\hat{B}_{t-1}$ and $\hat{B}_{t}$ to capture their similarities based on 3D box center, dimension, and rotation. 

We define each entry $(i,j)$ in $R_v$ as follows:
\begin{align}
    L_c(i,j) &= \Big|\Big|c_{t-1}^i - c_t^j \Big|\Big|_2^2 \label{eq:voxel_center}\\
    L_d(i,j) &= \Bigg|log\bigg(\frac{w_{t-1}^i}{w_t^j} \bigg)\Bigg| + \Bigg|log\bigg(\frac{l_{t-1}^i}{l_t^j} \bigg)\Bigg| + \Bigg|log\bigg(\frac{h_{t-1}^i}{h_t^j} \bigg)\Bigg| \label{eq:voxel_dim}\\
     L_r(i,j)^2 &= \Big(cos(r_{y, t-1}^i) - cos(r_{y, t}^j)\Big)^2 \nonumber\\
     &\quad\quad + \Big(sin(r_{y, t-1}^i) - sin(r_{y, t}^j)\Big)^2 \label{eq:voxel_rot}\\
    R_v(i,j) &= \hat{L}_c(i,j) + L_d(i,j) + L_r(i,j) \label{eq:voxel_res}
\end{align} 
where $c := (x,y,z)$ is the 3D box center. For the overall residual in Equation~\ref{eq:voxel_res}, $L_c(i,j)$ is normalized to make all the normalized elements $\hat{L}_c(i,j)$ dimensionless like the terms in $L_d$ and $L_r$. 

\noindent \textbf{Bounding Box Residual.} 
The learned bounding box residual $R_b \in \mathbb{R}^{({N_{max}+2}) \times ({N_{max}+2})}$ is found by first expanding $\hat{B}_{t-1}$ and $\hat{B}_t$ and concatenating them to get a matrix $\hat{B} \in \mathbb{R}^{(N_{max}+2) \times (N_{max}+2) \times 6}$. Note that for this step we only use the box centers from $\hat{B}_{t-1}$ and $\hat{B}_t$. Then, we obtain our residual $R_b \in \mathbb{R}^{(N_{max}+2) \times (N_{max}+2)}$ with an MLP $\sigma_r^b$ as such: 
\begin{align}
    R_b = \sigma_{r}^b(\hat{B}).
\end{align}

\noindent \textbf{Shape Residual.}
We use $\hat{S}_t$ and $\hat{S}_{t-1}$ to obtain the learned shape residual $R_s$ between the two frames. We expand $\hat{S}_{t-1}$ and $\hat{S}_t$ and concatenate them to get a matrix $\hat{S} \in \mathbb{R}^{(N_{max}+2)\times (N_{max}+2)\times 2F}$. Then, we obtain our residual $R_s \in \mathbb{R}^{(N_{max}+2)\times (N_{max}+2)}$ with an MLP: 
\begin{align}
    R_s = \sigma_r^s(\hat{S}).
\end{align}

\noindent \textbf{Overall Residual.}
The overall residual $R$ is a weighted sum of the three residuals we previously obtained: $R_v$, $R_b$, and $R_s$. We concatenate $\hat{B}$ and $\hat{S}$ that we previously obtained in the early steps to create our input $\hat{W} \in \mathbb{R}^{(N_{max}+2)\times(N_{max}+2)\times(2F+6)}$. Then, we pass this through an MLP $\sigma_\alpha$ to get the learned weights $\pmb{\alpha} \in \mathbb{R}^{(N_{max}+2)\times (N_{max}+2) \times 3}$ as follows: 
\begin{align}
    \pmb{\alpha} = \sigma_\alpha(\hat{W}).
\end{align}
We split $\pmb{\alpha}$ into matrices $\alpha_v$, $\alpha_b$, $\alpha_s \in \mathbb{R}^{(N_{max}+2)\times (N_{max}+2)}$, and obtain our overall residual $R \in \mathbb{R}^{(N_{max}+2)\times (N_{max}+2)}$:
\begin{align}
    R = \alpha_v \odot R_v + \alpha_b \odot R_b + \alpha_s \odot R_s.
\end{align}
Note that $\odot$ is the Hadamard product.

\subsubsection{Affinity Matrix Estimation}
Given our overall residual, we have information about the pairwise similarities between current frame detections and previous frame tracks, as well as each of the four augmented anchors. We want to use these spatio-temporal and shape relationships to learn probabilities for matching the detections and tracks to each other or the anchors. This will ultimately allow us to leverage global relationships for learning our matching probabilities, as opposed to using greedy local searches on residuals for matching~\cite{yin2021center, chiu2020probabilistic}. We apply an MLP to get our overall affinity matrix $A$:
\begin{align}
    A =  \sigma_{aff}(R).
\end{align}




\subsection{Log Affinity Loss}

Inspired by~\cite{guo2022joint, hayat2019max, dan2019}, we train our model with the log affinity loss $\mathcal{L}_{la}$. We define the ground-truth affinity matrix $A_{gt}$, the estimated affinity matrix $A$, and the Hadamard product $\odot$ to get:
\begin{align}
    \mathcal{L}_{la} = \frac{\sum_i \sum_j (A_{gt} \odot -\log(A))}{\sum_i \sum_j  A_{gt}}.
\end{align}



The affinity matrix prediction has corresponding ground-truth matrices $A_{gt,fm}$ and $A_{gt,bm}$ for forward matching and backward matching, respectively. In total, our overall loss function $\mathcal{L}$ is defined as follows:
\begin{align}
    \mathcal{L}_{fm} &= \frac{\sum_i \sum_j (A_{gt,fm} \odot -\log(A_{fm}))}{\sum_i \sum_j  A_{gt,fm}} \\ 
    \mathcal{L}_{bm} &= \frac{\sum_i \sum_j (A_{gt,bm} \odot -\log(A_{bm}))}{\sum_i \sum_j  A_{gt,bm}} \\
    \mathcal{L} &= \frac{1}{2} \Big(\mathcal{L}_{fm} + \mathcal{L}_{bm}\Big).
\end{align}

\subsection{Sequential Track Confidence Refinement}
We propose a sequential track confidence refinement technique that is used to assess the quality of tracks in real-time. We first offer an intuition behind our sequential track confidence refinement technique before formalizing our approach in Equation~\ref{eq:conf_ref}.

As stated earlier, track confidence should accurately reflect a track's quality relative to the quality of tracks for objects of the same class in a given time step. Thus, to obtain more accurate confidence scores to characterize our tracks, we want to leverage our affinity matrix estimation as follows: (1) If the affinity matrix indicates that the detection matched to our track has a high probability of being true-positive, we want to take a weighted average between the confidence of our matched detection and the existing confidence of our track. (2) If the affinity matrix indicates that the detection matched to our track almost got eliminated as an FP -- only marginally missing the elimination threshold $\tau_{FP}$ -- then we want to downscale the existing confidence for our track, because it likely should not be kept alive with this matched detection. This approach is captured in Equation~\ref{eq:conf_ref}. 
\begin{align}\label{eq:conf_ref}
    c_{trk, i}^{(t)} &\leftarrow \mathds{1}[P_{FP, i}^{(t)} < \beta_1]\beta_2 c_{det, i}^{(t)} + (1-\beta_2)c_{trk, i}^{(t-1)} \\
    \textrm{s.t.} \quad & 0 \leq c_{det,i}^{(t)} \leq 1 \quad\forall i,t \nonumber\\
    & 0 \leq c_{trk,i}^{(t)} \leq 1 \quad\forall i,t \nonumber\\
    & 0 \leq \beta_1 \leq 1 \nonumber\\
  & 0 \leq \beta_2 \leq 1 \nonumber
\end{align}

According to Equation~\ref{eq:conf_ref}, for track ID $i$ at time step $t$, we have a confidence of $c_{trk,i}^{(t)}$. The affinity matrix gets leveraged in the indicator function with $P_{FP, i}^{(t)}$, which is the probability that the detection got matched with the false-positive anchor at time step $t$. We fix $\beta_1$ to be a value slightly less than $\tau_{FP}$ to account for detections that are very close to getting eliminated as FP. In other words, we view detections with $P_{FP,i}^{(t)} \in [\beta_1, \tau_{FP}]$ to be very uncertain detections, and we take this into account by reducing the overall track confidence for tracks matching to such detections. However, if the matched detection has a very high probability of being true-positive, then the track confidence $c_{trk,i}^{(t)}$ becomes a weighted average between $c_{det,i}^{(t)}$ and $c_{trk,i}^{(t-1)}$. Based on cross validation, we found that this refinement technique is not very sensitive to values for $\beta_1$ and $\beta_2$. In fact, we set $\beta_1 = 0.5$ for all object classes, and $\beta_2=0.5$ for all object classes except for bicycle ($\beta_2=0.4$), bus ($\beta_2 =0.7$), and trailer ($\beta_2=0.4$). In the case of bicycle and trailer, we choose $\beta$ to be slightly lower, because we want to give more weight to the fact that it has been tracked since 3D detectors tend not to be very good on these object types and thus skew towards lower detection confidence scores. For the converse reason, we give bus a higher $\beta_2$.

In the special case of a newborn track, the track confidence is only the first term of Equation~\ref{eq:conf_ref}, i.e. $$c_{trk,i}^{(t)} \leftarrow \mathds{1}[P_{FP, i}^{(t)} < \beta_1]\beta_2 c_{det, i}^{(t)}.$$ This works well, because the detection confidences are also created in relative terms. Thus, if all newly initialized track confidences are the detection confidences scaled with $\beta_2$, their relative confidence rankings will be preserved. 



\section{Data Preparation}
We evaluate our tracking performance on the nuScenes benchmark~\cite{caesar2020nuscenes}. The nuScenes dataset contains 1000 scenes, where each scene is 20 seconds long and is generated with a 2Hz frame-rate. For the tracking task, nuScenes has 7 classes: bicycle, bus, car, motorcycle, pedestrian, trailer, and truck. In this section, we provide an overview of the LiDAR preprocessing and ground-truth affinity matrix formation we complete to deploy ShaSTA for the nuScenes dataset.

\begin{table*}[th!]
\caption{nuScenes Test Results with evaluation in terms of overall and individual AMOTA and AMOTP. Our method achieves the highest AMOTA, while effectively balancing AMOTP to get the second lowest AMOTP only 0.005m shy of the first spot. We only compare on LiDAR-only methods that use CenterPoint detections to ensure a fair comparison. The blue entry is our LiDAR-only method ShaSTA.}
\vspace{-10pt}
\label{table_nusc1}
\begin{center}
\begin{tabular}{c|c|c|c|c|c|c|c|c|c|c|c }
Metric & Method & Detector & Input & Overall & Bicycle & Bus & Car & Motorcycle & Pedestrian & Trailer & Truck \\
\hline
\multirow{5}{*}{AMOTA $\uparrow$}
& \cellcolor{blue!15}ShaSTA (Ours)& \cellcolor{blue!15}CenterPoint & \cellcolor{blue!15}LiDAR & \cellcolor{blue!15}\textbf{69.6} & \cellcolor{blue!15}41.0 & \cellcolor{blue!15}\textbf{73.3} & \cellcolor{blue!15}\textbf{83.8} & \cellcolor{blue!15}\textbf{72.7} & \cellcolor{blue!15}\textbf{81.0} & \cellcolor{blue!15}\textbf{70.4} & \cellcolor{blue!15}\textbf{65.0} \\
& NEBP~\cite{liang2022neural} & CenterPoint & LiDAR & 68.3 & \textbf{44.7} & 70.8 & 83.5 & 69.8 & 80.2 & 69.0 & 59.8 \\
& OGR3MOT~\cite{zaech2022learnable} & CenterPoint & LiDAR & 65.6 & 38.0 & 71.1 & 81.6 & 64.0 & 78.7 & 67.1 & 59.0 \\
& CenterPoint~\cite{yin2021center} & CenterPoint & LiDAR & 65.0 & 33.1 & 71.5 & 81.8 & 58.7 & 78.0 & \textbf{69.3} & 62.5 \\
\hline
\multirow{5}{*}{AMOTP $\downarrow$}
& \cellcolor{blue!15}ShaSTA (Ours)& \cellcolor{blue!15}CenterPoint & \cellcolor{blue!15}LiDAR & \cellcolor{blue!15}0.540 &  \cellcolor{blue!15}0.674 & \cellcolor{blue!15}\textbf{0.629} & \cellcolor{blue!15}\textbf{0.383} & \cellcolor{blue!15}\textbf{0.504} & \cellcolor{blue!15}\textbf{0.369} & \cellcolor{blue!15}0.742 & 
\cellcolor{blue!15}0.650\\
& NEBP~\cite{liang2022neural} & CenterPoint & LiDAR & 0.624 & 1.026 & 0.677 & 0.391 & 0.557 & 0.393 & 0.781 & 0.541 \\
& OGR3MOT~\cite{zaech2022learnable} & CenterPoint & LiDAR & 0.620 & 0.899 & 0.675 & 0.395 & 0.615 & 0.383 & 0.790 & 0.585 \\
& CenterPoint~\cite{yin2021center} & CenterPoint & LiDAR & \textbf{0.535} & \textbf{0.561} & 0.636 & 0.391 & 0.519 & 0.409 & \textbf{0.729} & \textbf{0.5} \\
\hline
\end{tabular}
\end{center}
\end{table*}

\begin{table}[h!]
\caption{nuScenes Test Results with evaluation in terms of overall TP, FP, and FN. We only compare on LiDAR-only methods that use CenterPoint detections to ensure a fair comparison. The blue entry is our LiDAR-only method ShaSTA.}
\vspace{-10pt}
\label{table_nusc2}
\begin{center}
\begin{tabular}{c |c | c | c c c}
\hline
Method & Detector & Input & TP $\uparrow$ & FP $\downarrow$ & FN $\downarrow$\\
\hline 
\cellcolor{blue!15}ShaSTA (Ours)& \cellcolor{blue!15}CenterPoint & \cellcolor{blue!15}LiDAR & \cellcolor{blue!15}\textbf{97,799} & \cellcolor{blue!15}\textbf{16,746} & \cellcolor{blue!15}\textbf{21,293} \\
\hline
NEBP~\cite{liang2022neural} & CenterPoint & LiDAR & 97,367 & 16,773 & 21,971 \\
\hline
OGR3MOT~\cite{zaech2022learnable} & CenterPoint & LiDAR & 95,264 & 17,877 & 24,013 \\
\hline 
CenterPoint~\cite{yin2021center} & CenterPoint & LiDAR & 94,324 & 17,355 & 24,557 \\
\hline
\end{tabular}
\end{center}
\end{table}

\subsection{LiDAR Point Cloud Preprocessing}
Since CenterPoint~\cite{yin2021center} provides detections at 2Hz, but the LiDAR has a sampling rate of 20Hz, we use the nuScenes~\cite{caesar2020nuscenes} feature for accumulating multiple LiDAR sweeps with motion compensation. This provides us with a denser 4D point cloud with an added temporal dimension, and makes the LiDAR input match the detection sampling rate. We use 10 LiDAR sweeps to generate the point cloud.

\subsection{Ground-Truth Affinity Matrix.} We find the true-positive, false-positive, and false-negative detections using the matching algorithm from the nuScenes~\cite{caesar2020nuscenes} dev-kit. For each frame, we define a ground-truth affinity matrix $A_t \in \mathbb{R}^{(N_{max}+2) \times (N_{max}+2)}$ between the current frame and previous frame off-the-shelf detections. For all true-positive previous and current frame detections, $B_{t-1}^{tp}$ and $B_t^{tp}$, we say they are matched if the ground-truth boxes they each correspond to have the same ground-truth tracking IDs. If there is a match between true-positive boxes $b_{t-1, i}^{tp} \in B_{t-1}^{tp}$ and $b_{t, j}^{tp} \in B_t^{tp}$, then $A_t(i,j) = 1$. For all false-positive previous frame detections and true-positive previous frame detections whose tracking IDs do not appear in any of the current frame ground-truth tracking IDs, we call them dead tracks $B_{t-1}^{dt}$. For all $b_{t-1, i}^{dt} \in B_{t-1}^{dt}$, we set $A_t(i,N_{max}+1) = 1$. For all true-positive previous frame detections that do not have a match in the current frame but whose tracking IDs do appear in the current frame ground-truth tracking IDs, we call them false-negatives $B_{t-1}^{fn}$. For all $b_{t-1, i}^{fn} \in B_{t-1}^{fn}$, we set $A_t(i,N_{max}+2) = 1$. Additionally, for all true-positive current frame detections whose tracking IDs do not appear in the previous frame ground-truth tracking IDs, we call them newborn tracks $B_{t}^{nb}$, and we set $A_t(N_{max}+1, j) = 1$ for all $b_{t, j}^{nb} \in B_{t}^{nb}$. Finally, we define the set of false-positive current frame detections as $B_{t}^{fp}$, and we set $A_t(N_{max}+2, j) = 1$ for all $b_{t, j}^{fp} \in B_{t}^{fp}$. Unmatched $(i,j)$ entries are set to $0$.


\section{Experimental Results}\label{sec:experiments}

This section provides an overview of evaluation metrics, training details, comparisons against state-of-the-art 3D multi-object tracking techniques, and ablation studies to analyze our technique. 

\subsection{Evaluation Metrics}\label{sec:metrics}
The primary metrics for evaluating nuScenes are Average Multi-Object Tracking Accuracy (AMOTA) and Average Multi-Object Tracking Precision (AMOTP)~\cite{weng2019baseline}. AMOTA measures the ability of a tracker to track objects correctly, while AMOTP measures the quality of the estimated tracks. 

AMOTA averages the recall-weighted MOTA, known as MOTAR, at $n$ evenly spaced recall thresholds. Note that MOTA is a metric that penalizes FPs, FNs, and ID switches (IDS). GT indicates the number of ground-truth tracklets.
{\small
\begin{align}
    AMOTA &= \frac{1}{n-1} \sum_{r \in \{\frac{1}{n-1},\frac{1}{n-2},\dots,1\}} MOTAR \\
    MOTAR &= max\Bigg(0, 1 - \frac{IDS_r + FP_r + FN_r - (1-r)\cdot GT}{r \cdot GT}\Bigg)
\end{align}}
Additionally AMOTP averages MOTP over $n$ evenly spaced recall thresholds. Note that MOTP measures the misalignment between ground-truth and predicted bounding boxes. For the definition of AMOTP below, $d_{i,t}$ indicates the distance error for track $i$ at time step $t$ and $TP_t$ indicates the number of true-positive (TP) matches at time step $t$.
\begin{align}
    AMOTP = \frac{1}{n-1} \sum_{r \in \{\frac{1}{n-1},\frac{1}{n-2},\dots,1\}} \frac{\sum_{i,t}d_{i,t}}{\sum_t TP_t} 
\end{align}
The nuScenes benchmark~\cite{caesar2020nuscenes} defines TP tracks as estimated tracks that are within an L2 distance of 2m from ground-truth tracks. This allows for a margin of error that can become unsafe in certain driving scenarios where small distances can make a dramatic difference in safety. Thus, even though the official nuScenes leaderboard uses AMOTA solely to rank trackers, we emphasize AMOTP as well due to its safety implications.

 We also present secondary metrics, including the total number of TP, FP, and FN tracks to better analyze the effectiveness of our approach for downstream decision-making tasks.

\begin{table*}[ht!]
\caption{\textbf{Ablation on Confidence Refinement.} We assess the effect of confidence refinement on tracking performance for each object. The version of ShaSTA without refinement indicates that we apply the CenterPoint detection confidence scores to our tracks. Evaluation is on nuScenes validation set in terms of overall AMOTA. The blue entry is our full method. }
\vspace{-10pt}
\label{conf_ref_ablation}
\begin{center}
\begin{tabular}{c|c|c|c|c|c|c|c|c|c}
\hline
Metric & Method & Overall & Bicycle & Bus & Car & Motorcycle & Pedestrian & Trailer & Truck \\
\hline
\multirow{2}{*}{AMOTA $\uparrow$}
& Without Refinement & 70.2 & 49.4 & 85.2 & 85.3 & 70.1 & 80.9 & 50.1 & 70.1 \\
& \cellcolor{blue!15} With Refinement & \cellcolor{blue!15}\textbf{72.8} & \cellcolor{blue!15}\textbf{58.8} & \cellcolor{blue!15}\textbf{85.6} & \cellcolor{blue!15}\textbf{85.6} & \cellcolor{blue!15}\textbf{74.5} & \cellcolor{blue!15}\textbf{81.4} & \cellcolor{blue!15}\textbf{53.1} & \cellcolor{blue!15}\textbf{70.3} \\
\hline
& \textit{Difference} & \textit{+2.6} & \textit{+9.4} & \textit{+0.4} & \textit{+0.3} & \textit{+4.4} & \textit{+0.5} & \textit{+3.0} & \textit{+0.2} 
 \\
\hline
\end{tabular}
\end{center}
\end{table*}

\subsection{Training Specifications}\label{sec:training}
We train a different network for each object category following~\cite{chiu2021probabilistic, stearns2022spot}. The value of $N_{max}$ depends on the object type since some classes are more common than others. We choose the value for $N_{max}$ based on the per-object detection frequency we gather in the training set. The smallest is $N_{max} = 20$ and the largest is $N_{max} = 90$. We use the Adam optimizer with a constant learning rate of $10^{-4}$ and L2 weight regularization of $10^{-2}$. We use the pre-trained VoxelNet weights from CenterPoint~\cite{yin2021center} and freeze them for our LiDAR backbone so that the shape information correlates with our detection bounding boxes from \cite{yin2021center}. Additionally, there is a significant class imbalance between FPs and TPs since state-of-the-art detectors function in low precision-high recall regimes.
Thus, we downsample the number of FP detections during training.
Finally, we fix the thresholds $\tau_{fp}$ for FP elimination, $\tau_{fn}$ for FN propagation, $\tau_{nb}$ for NB initialization, and $\tau_{dt}$ for DT termination with cross-validation. We found that $\tau_{fp}=0.7$, $\tau_{fn}=0.5$, $\tau_{nb}=0.5$, and $\tau_{dt}=0.5$ works best for all classes.



\subsection{Comparison with State-of-the-Art LiDAR Tracking} Our nuScenes test results can be seen in Tables~\ref{table_nusc1} and~\ref{table_nusc2}. Because previous works have shown that the standard AMOTA tracking metric heavily depends on the upstream 3D detections~\cite{chiu2021probabilistic, luiten2021hota, wang2022camo}, all tracking methods reported use CenterPoint detections to ensure a fair comparison\footnote{At the time of this writing, CenterPoint detections achieve the highest mAP of all publicly-available 3D detections. Higher-scoring detections appear on the nuScenes leaderboard, but are proprietary.}. For more information on this evaluation protocol and the AMOTA metric, please refer to Section~\ref{sec:metrics}.




The CenterPoint tracking algorithm~\cite{yin2021center} is our baseline, and we include recent, peer-reviewed works cited in Section~\ref{sec:related} that tackle data association, track lifecycle management, FPs, and FNs in the LiDAR-only tracking domain. Most notably, before ShaSTA was developed, NEBP~\cite{liang2022neural} was the \#1 LiDAR-only tracker using CenterPoint detections.

In Table~\ref{table_nusc1}, we can see that ShaSTA balances correct tracking (AMOTA) with precise, high-quality tracking (AMOTP). Most notably ShaSTA not only achieves the highest overall AMOTA, but also does so for 6 out of 7 object classes. The only class we are not first in AMOTA is the bicycle class, with NEBP taking the lead. However, it is interesting to note that our precision is much better for bicycles than NEBP is. NEBP averages over 1 meter in distance from the ground-truth tracks on bicycles, which can be very dangerous in real-world traffic scenes. Compared to NEBP, ShaSTA has a 34\% improvement in AMOTP while only having an 8\% reduction in AMOTA. Thus, the trade-off for the bicycles is clearly in ShaSTA's favor. For the overall AMOTP, we achieve the second best value only trailing CenterPoint by 0.005 meters, which is a favorable trade-off when compared to our significant boost in overall AMOTA. 

Examining further metric breakdowns in Table~\ref{table_nusc2}, it becomes clear that ShaSTA is effective in de-cluttering environments suffering from FPs and compensating for missed detections that lead to missing TPs and rising FNs. The two most observed classes that constitute over 80\% of the test set are cars and pedestrians, and we get highly competitive results in both of those categories, which tend to suffer the most from crowded scenes. Thus, our FPs reflect our effectiveness at dealing with these sorts of scenes. Another interesting interplay between these metrics is that often times decreasing FPs comes at the risk of increasing FNs and decreasing TPs. Thus, it is quite challenging to tackle the FP, FN, and TP problem simultaneously, yet our approach does so most effectively out of all existing tracking methods. 



\begin{table}[h]
\caption{\textbf{Ablation on Shape Descriptors.} We assess the effect of the number of 3D bounding box points used to extract shape descriptors. Evaluation is on nuScenes validation set in terms of overall AMOTA. The blue entry is our full method. }
\vspace{-10pt}
\label{shape_ablation}
\begin{center}
\begin{tabular}{c c|c }
\hline
Box Center Used & Face Centers Used & AMOTA $\uparrow$\\
\hline 
\checkmark & \xmark & 68.8\\ 
\hline
\xmark & \checkmark &  70.2\\
\hline 
\cellcolor{blue!15}\checkmark & \cellcolor{blue!15}\checkmark & \cellcolor{blue!15}\textbf{72.8} \\
\hline 
\end{tabular}
\end{center}
\end{table}

\begin{table}[h]
\caption{\textbf{Ablation on Residuals.} We assess the effect of isolating each residual on the tracking algorithm's performance. Evaluation is on nuScenes validation set in terms of overall AMOTA. The blue entry is our full method. }
\vspace{-10pt}
\label{residual_ablation}
\begin{center}
\begin{tabular}{c|c}
\hline
Residual Used & AMOTA $\uparrow$\\ 
\hline 
VoxelNet Only & 68.9\\ 
\hline
Bounding Box Only & 69.1\\ 
\hline 
Shape Only & 69.5\\ 
\hline 
\cellcolor{blue!15}All (ShaSTA) & \cellcolor{blue!15}\textbf{72.8} \\
\hline 
\end{tabular}
\end{center}
\end{table}

\begin{table}[h]
\caption{\textbf{Ablation on Learned Affinity Matrix.} We assess the effect of each augmentation from our learned affinity matrix on track formation. Evaluation is on nuScenes validation set in terms of overall AMOTA. The blue entry is our full method. }
\vspace{-10pt}
\label{aug_ablation}
\begin{center}
\begin{tabular}{c |c c}
\hline
Affinity Augmentation Method & AMOTA $\uparrow$\\ 
\hline 
False-Positive Elimination Only & 72.7\\
\hline
False-Negative Propagation Only & 70.4\\
\hline 
Newborn Initialization Only & 72.6\\
\hline
Dead Track Termination Only & 70.4\\
\hline
\cellcolor{blue!15}All (ShaSTA) & \cellcolor{blue!15}\textbf{72.8}\\
\hline
\end{tabular}
\end{center}
\end{table}


\subsection{Ablation Studies}
We complete four ablative experiments on the nuScenes validation set to measure the impacts of (1) the sequential track confidence refinement, (2)  the number of shape descriptors from the BEV map, (3) each of the residuals used to make the overall residual prediction, and (4) including FP elimination, FN propagation, NB initialization, and DT termination in the learned affinity matrix. In total, our ablative studies demonstrate that all of the components in ShaSTA contribute to the overall success of this tracking framework. 

\begin{figure*}[ht!]
    \centering
    \begin{subfigure}[b]{\textwidth}
        \centering
        \includegraphics[width=\textwidth]{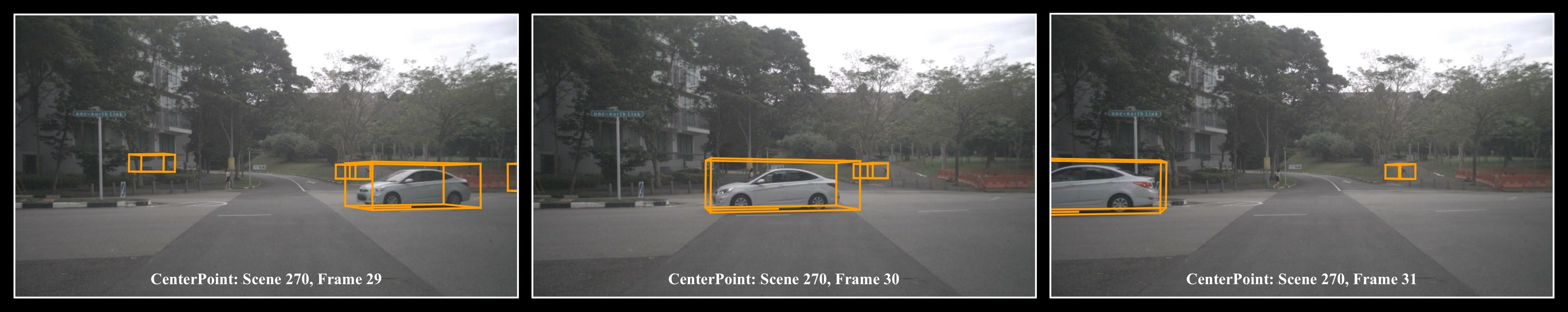}
        \caption{CenterPoint: Scene 270}   
        \label{fig:cp_29}
    \end{subfigure}
    \vskip\baselineskip
    \vspace{-5pt}
    \begin{subfigure}[b]{\textwidth}   
        \centering 
        \includegraphics[width=\textwidth]{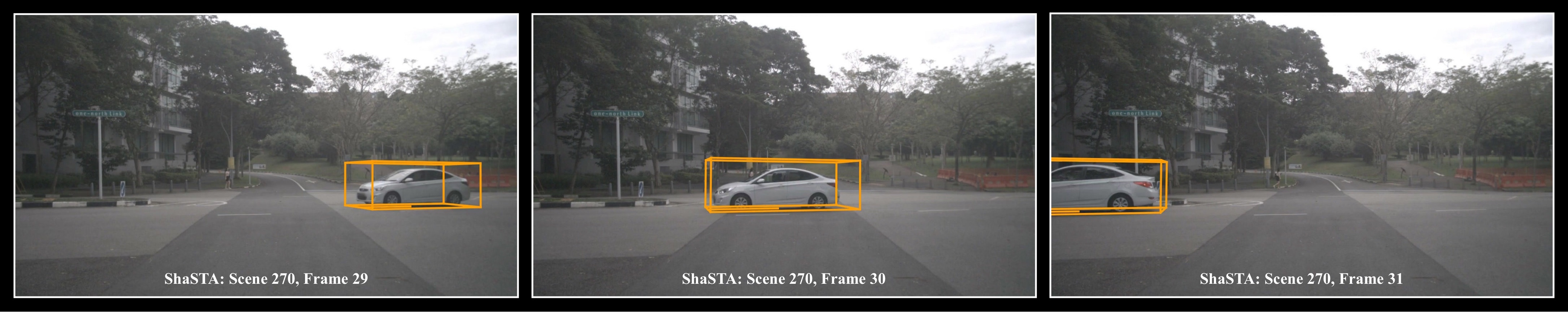}
        \caption{ShaSTA: Scene 270}    
        \label{fig:shasta_29}
    \end{subfigure}
    \caption{Visualization of tracking results for car object class projected onto front camera. We show two consecutive frames from Scene 270 of the validation set for CenterPoint~\cite{yin2021center} and ShaSTA. In frame 29, CenterPoint has 3 false-positive tracks as can be seen in frame 29, and one of the false-positives persists into frames 30 and 31. However, ShaSTA perfectly tracks the car in this sequence without any false-positives or false-negatives. Note that both methods are starting with the CenterPoint detections as input.} 
    \label{fig:qual_res}
\end{figure*}

First, we analyze the effect of the sequential tracking confidence refinement as seen in Table~\ref{conf_ref_ablation}. When we do not use our refinement technique, we apply the CenterPoint detection confidence scores to our tracks as in past works~\cite{weng2019ab3dmot,liang2022neural,chiu2020probabilistic,chiu2021probabilistic,yin2021center,zaech2022learnable}. We note a significant boost in the overall AMOTA for the validation set when using the proposed refinement method. All object classes benefited, but the ones that improved the least are large objects like trucks, cars, and buses. This makes sense since these types of objects tend to provide the best LiDAR point clouds and are less likely to suffer from full occlusion, allowing the detection confidence to be a good indicator for tracking. Conversely, smaller objects like bicycles and motorcycles that tend to suffer the most from LiDAR point cloud sparsity due to increased distance from the ego vehicle and occlusion in traffic scenes benefit the most from this technique. In these cases, the pure detection confidences can drop dramatically when there are scenarios such as full occlusion of these small objects, but with our sequential refinement we take into account the existing confidence that has accumulated for a track before such scenarios occur. 

Additionally, in Table~\ref{shape_ablation}, we can see that as we add more bounding box face centers for extracting shape descriptors, the accuracy of our model monotonically increases. This result supports our argument that leveraging raw LiDAR point clouds to get shape information is integral to the success of our tracking framework. Most notably, when we have less shape information, more objects within the same class are more likely to appear similar, leading to more ambiguity in distinguishing between TPs and FPs. Such ambiguity makes the track confidence refinement less effective, specifically in the indicator function in Equation~\ref{eq:conf_ref}.

Furthermore, our ablation on the individual residuals in Table~\ref{residual_ablation} demonstrates that even though we can achieve competitive tracking accuracy solely with spatio-temporal (VoxelNet or Bounding Box only) or shape (Shape only) information, we achieve the most optimal results when leveraging both types of information. 

Lastly, in Table~\ref{aug_ablation}, we see that the most impactful augmentations are FP elimination and NB initialization. Eliminating FPs is expected to boost our accuracy since off-the-shelf detections suffer from such high rates of FP detections. NB initialization can be seen as a second reinforcement on eliminating FP detections because unmatched detections that do not have a probability of at least $\tau_{nb}$ for being NBs do not get initialized as tracks and are discarded. Finally, even though FNs and DT termination have not been as detrimental to tracking algorithms in the past, these augmentations alone give us competitive accuracies, showing that every augmentation contributes to effective tracking.

\subsection{Qualitative Results}
As indicated in Table~\ref{table_nusc1}, there are significant increases in test accuracy over the CenterPoint~\cite{yin2021center} baseline, particularly for objects like cars and pedestrians that are often found in crowded scenes and thus, have increased susceptibility to experiencing high rates of FP tracks.

At high recall thresholds, FP tracks tend to appear more frequently, but the AMOTA metric does not strongly penalize FP tracks at higher recall thresholds, diminishing their significant effect on real-world deployment for downstream decision-making tasks. To showcase ShaSTA's ability to eliminate FPs effectively, we present tracking results in Figure~\ref{fig:qual_res}. We only show the tracking results on the car object class from validation scene 270. (a) shows the CenterPoint~\cite{yin2021center} baseline which has FP tracks in all three frames. On the other hand, (b) demonstrates ShaSTA's ability to declutter the scene and accurately track the car, while ignoring background information that should not be tracked. Typically, past works~\cite{stearns2022spot} use non-maximal suppression heuristics such as intersection over union (IoU) to remove FP tracks, but in more complex settings like scene 270, such heuristics would fail because none of the FPs overlap with other track predictions. As a result, ShaSTA proves to be especially effective at using shape and spatio-temporal information from LiDAR point clouds to localize and track actual objects of interest in the scene.
\section{Conclusion}
We present ShaSTA, a 3D multi-object tracking framework that leverages shape and spatio-temporal information from LiDAR sensors to learn affinities for robust data association, track lifecycle management, FP elimination, FN propagation, and track confidence refinement.
Our approach effectively addresses false-positives and missing detections against cluttered scenes and occlusion, yielding state-of-the-art performance. 
Since ShaSTA offers a flexible framework, we can easily extend it in future work to fuse various sensor modalities. Other interesting future directions include the usage of our learned affinities in prediction and planning.
\section{Acknowledgment}
Toyota Research Institute provided funds to support this work.

{
    \footnotesize
    \bibliographystyle{IEEEtran}
    \bibliography{IEEEabrv, refs.bib}
}

\end{document}